%% file: iclr2025_conference.tex
\documentclass{article} %
\usepackage{iclr2025_conference,times}

\input{math_commands.tex}

\usepackage{hyperref}
\usepackage{url}

\hypersetup{
    colorlinks = true,
    linkcolor  = red,
    urlcolor   = magenta,
    citecolor  = teal,
}

\usepackage[utf8]{inputenc} %
\usepackage[T1]{fontenc}    %
\usepackage{hyperref}       %
\usepackage{url}            %
\usepackage{booktabs}       %
\usepackage{amsfonts}       %
\usepackage{nicefrac}       %
\usepackage{microtype}      %
\usepackage{xcolor}         %
\usepackage{enumitem}
\usepackage{multirow}
\usepackage{pifont}
\usepackage{graphicx}
\usepackage{amsmath}
\usepackage{algorithm}
\usepackage{mathtools}
\usepackage{algpseudocode}
\usepackage{listings}
\usepackage{wrapfig}
\usepackage{subcaption}
\usepackage{xcolor,colortbl}
\usepackage{array}

\newcommand{\Mat}{\boldsymbol}

\newcommand{\real}{\mathbb{R}}

\newcommand{\ord}[1]{#1^{\rm{th}}}

\definecolor{mygray}{gray}{0.94}

\definecolor{MyGreen}{rgb}{0.02,0.7,0.02}

\title{4K4DGen:\\ Panoramic 4D Generation at 4K Resolution}

\author{Renjie Li$\bf ^1$, Panwang Pan$\bf ^1$\thanks{Corresponding Author}~,  Bangbang Yang$\bf ^1$,  Dejia Xu$\bf ^2$,  Shijie Zhou$\bf ^3$, Xuanyang Zhang$\bf ^1$,\\  \textbf{Zeming Li$\bf ^1$,  Achuta Kadambi$\bf ^3$,  Zhangyang Wang$\bf ^2$,  Zhengzhong Tu$\bf ^4$,  Zhiwen Fan$\bf ^2$} \\
$^1$Pico, $^2$ UT Austin, $^3$ UCLA, $^4$ TAMU\\
\texttt{paulpanwang@gmail.com}
}

\iclrfinalcopy %
\begin{document}

\maketitle

\begin{figure}[ht]
  \centering
  \includegraphics[width=1\textwidth]{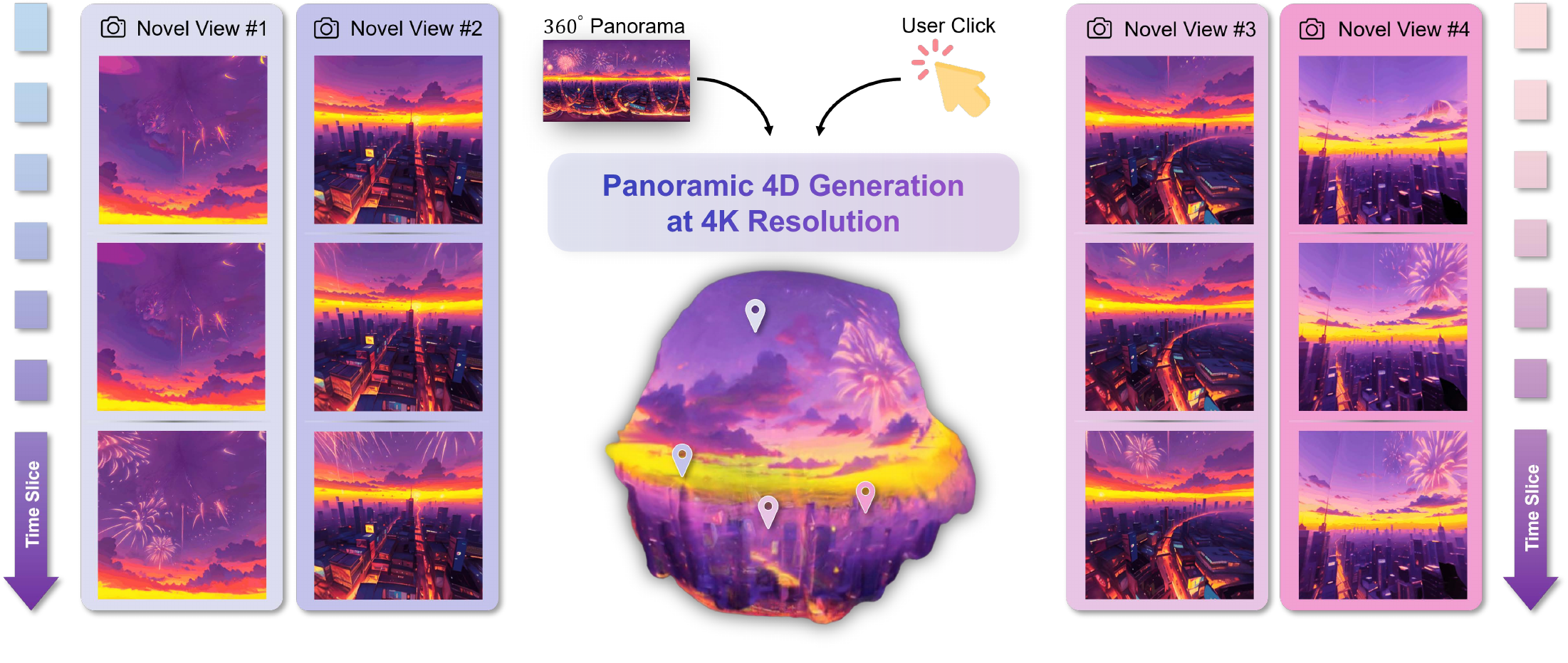}
\caption{\textbf{4K4DGen} takes a static panoramic image with a resolution of 4096$\times$2048 and allows animation through user interaction or an input mask, transforming the static panorama into dynamic Gaussian Splatting. 4K4DGen supports the rendering of novel views at various timestamps, enriching immersive virtual exploration.}
  \label{fig:teaser}
\end{figure}

\begin{abstract}
The blooming of virtual reality and augmented reality (VR/AR) technologies has driven an increasing demand for the creation of high-quality, immersive, and dynamic environments. However, existing generative techniques either focus solely on dynamic objects or perform outpainting from a single perspective image, failing to meet the requirements of VR/AR applications that need free-viewpoint, 360$^{\circ}$ virtual views where users can move in all directions. In this work, we tackle the challenging task of elevating a single panorama to an immersive 4D experience. For the first time, we demonstrate the capability to generate omnidirectional dynamic scenes with 360$^{\circ}$ views at 4K (4096 $\times$ 2048) resolution, thereby providing an immersive user experience. Our method introduces a pipeline that facilitates natural scene animations and optimizes a set of dynamic Gaussians using efficient splatting techniques for real-time exploration. To overcome the lack of scene-scale annotated 4D data and models, especially in panoramic formats, we propose a novel \textbf{Panoramic Denoiser} that adapts generic 2D diffusion priors to animate consistently in 360$^{\circ}$ images, transforming them into panoramic videos with dynamic scenes at targeted regions. Subsequently, we propose \textbf{Dynamic Panoramic Lifting} to elevate the panoramic video into a 4D immersive environment while preserving spatial and temporal consistency. By transferring prior knowledge from 2D models in the perspective domain to the panoramic domain and the 4D lifting with spatial appearance and geometry regularization, we achieve high-quality Panorama-to-4D generation at a resolution of 4K for the first time. 
\end{abstract}

\input{sec/intro}
\input{sec/related}

\input{sec/method}

\input{sec/exps}

\input{sec/conclusion}

\bibliography{iclr2025_conference}
\bibliographystyle{iclr2025_conference}

\appendix
\input{sec/supp}

\end{document}

%% file: math_commands.tex
\usepackage{amsmath,amsfonts,bm}

\def\eqref#1{equation~\ref{#1}}

\def\1{\bm{1}}

\DeclareMathAlphabet{\mathsfit}{\encodingdefault}{\sfdefault}{m}{sl}
\SetMathAlphabet{\mathsfit}{bold}{\encodingdefault}{\sfdefault}{bx}{n}

%% file: sec/intro.tex
\section{Introduction}
With the increasing growth of generative techniques~\citep{rombach2022high,blattmann2023stable}, the capability to create high-quality assets has the potential to revolutionize content creation across VR/AR and other spatial computing platforms. Unlike 2D displays such as smartphones or tablets, 
ideal VR/AR content must deliver an immersive and seamless experience, enabling 6-DoF virtual tours and supporting high-resolution 4D environments with omnidirectional viewing capabilities.
Despite significant advancements in the generation of images, videos, and 3D models, the development of panoramic 4D content has lagged, primarily due to the scarcity of well-annotated, high-quality 4D training data.
Even in the most relevant field of 4D generation, existing works mainly focus on generating or compositing object-level contents~\citep{bahmani20234d,lin2024instructlayout}, which are often in low-resolution (e.g., below 1080p) and cannot fulfill the demand of qualified immersive experiences.
Based on these observations, we propose that an ideal generative tool for creating immersive environments should possess the following properties: 
\textbf{(i)} the generated content should exhibit high perceptual quality, reaching high-resolution (4K) output with dynamic elements (4D);
\textbf{(ii)} the 4D representation must be capable of rendering coherent, continuous, and seamless 360$^{\circ}$ panoramic views in real time, supporting efficient 6-DoF virtual tours.
However, creating diverse, high-quality 4D panoramic assets presents two significant challenges:
\textbf{(i)} the scarcity of large-scale, annotated 4D data, particularly in panoramic formats, limits the training of specialized models.
\textbf{(ii)} achieving both fine-grained local details and global coherence in 4D and 4K panoramic views is difficult for existing 2D diffusion models. These models, typically trained on perspective images with narrow fields of view (FoV), cannot be easily adapted to the expansive scopes of large panoramic images (see Exp.~\ref{ssec:exp_ablation}).
On another front, video diffusion models \citep{an2023latent} trained with web-scale multi-modal data have demonstrated versatile utility as region-based dynamic priors, and Gaussian Splatting~\citep{kerbl20233d} has shown efficient capabilities in modeling 4D environment.
Thus, we address the large-scale, omnidirectional dynamic scene generation (4D panoramic generation) problem by utilizing the generative power of diffusion models to animate static panoramic images, transforming them into realistic, dynamic scenes that can support immersive, 360$^\circ$ viewing experiences. To achieve this, we propose to elevate the dynamic panoramic video to 4D environment assets using a set of dynamic Gaussians, which can be seamlessly integrated into VR/AR platforms for real-time rendering and interaction.

In this paper, we introduce \textbf{4K4DGen}, a novel framework designed to enable the creation of panoramic 4D environments at resolutions up to 4K. 4K4DGen addresses the key challenges of maintaining consistent object dynamics across the entire 360$^{\circ}$ field-of-view (FoV) in panoramic videos, while preserving both spatial and temporal coherence as the video transitions into a fully interactive 4D environment.
Specifically, we propose the \textbf{Panoramic Denoiser}, which animates 360$^{\circ}$ FoV panoramic images by denoising spherical latent codes corresponding to user-interacted regions. The Panoramic Denoiser leverages a well-trained diffusion model originally designed for narrow-FoV perspective images, enabling the generation of 360$^\circ$ dynamic panoramas while ensuring global coherence and continuity throughout the entire panorama.
To transform the omnidirectional panoramic video into a 4D environment, we introduce \textbf{Dynamic Panoramic Lifting}, which corrects scale discrepancies using a depth estimator enriched with perspective prior knowledge to generate panoramic depth maps. Additionally, it employs time-dependent 3D Gaussians optimized with spatial-temporal geometry alignment to ensure cross-frame consistency in dynamic scene representation and rendering.
By adapting generic 2D statistical patterns from the perspective domain to the panoramic format and effectively regularizing Gaussian optimization with geometric principles, we achieve high-quality 4K panorama-to-4D content generation with photorealistic novel-view synthesis capabilities.
Our contributions can be summarized as follows.
\begin{itemize}
\item We introduce \textbf{4K4DGen}, the first framework capable of generating high-resolution (up to 4096$\times$2048) 4D omnidirectional assets without the need for annotated 4D data. 
\item We propose the \textbf{Panoramic Denoiser}, which transfers generative priors from pre-trained 2D perspective diffusion models to the panoramic space, enabling consistent animation of panoramas with dynamic scene elements. 
\item We introduce \textbf{Dynamic Panoramic Lifting}, a method that transforms dynamic panoramic videos into dynamic Gaussians, incorporating spatial-temporal regularization to ensure cross-frame consistency and coherence. 
\end{itemize}

%% file: sec/related.tex
\begin{figure}[ht]
  \centering
  \vspace{-1em}
  \includegraphics[width=0.75\textwidth]{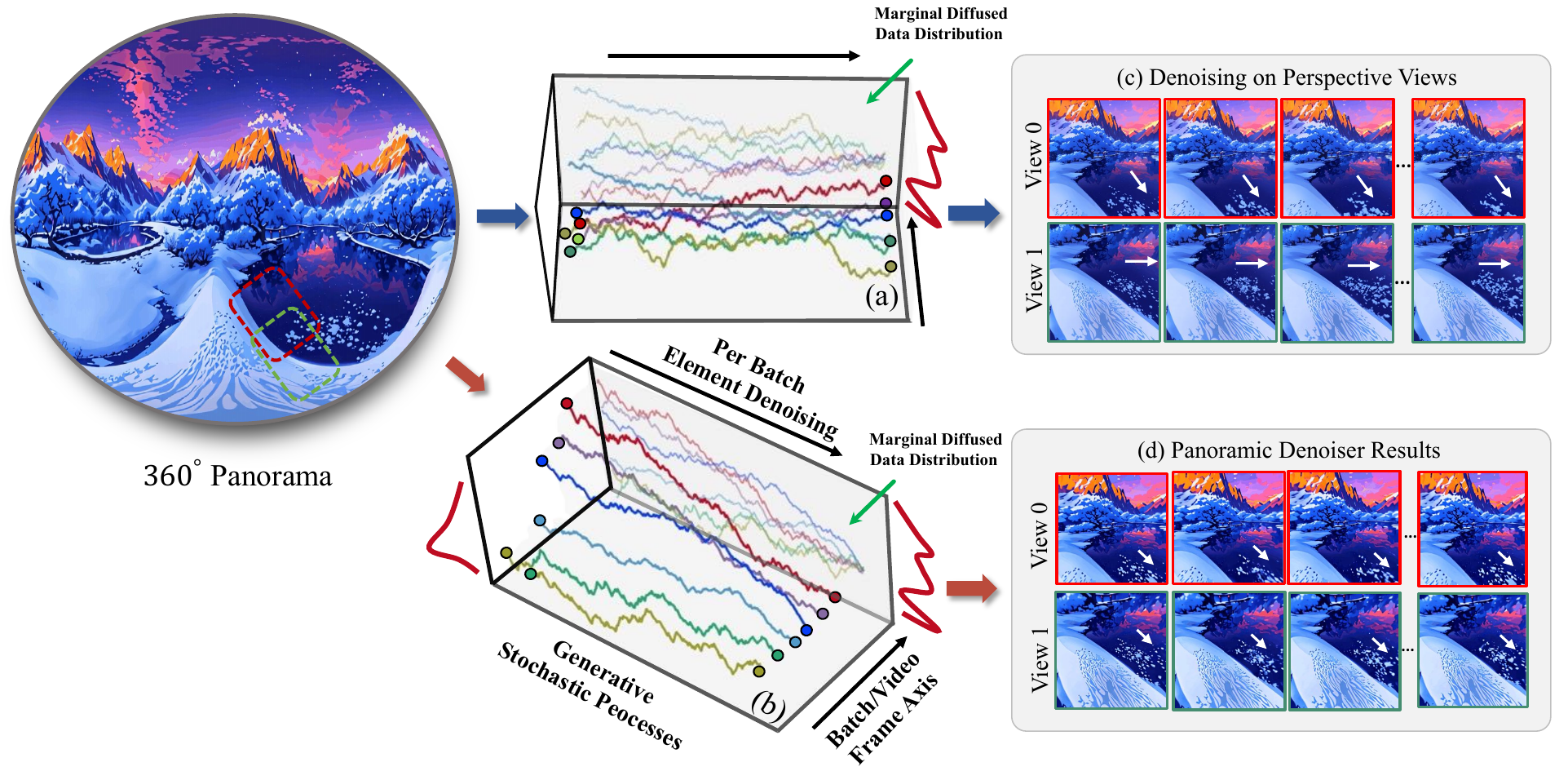}
\caption{\textbf{Panoramic Denoiser} adapts diffusion priors from the perspective domain to the panoramic domain by simultaneously denoising perspective views and integrating them into spherical latents at each denoising step. This approach ensures consistent animation across multiple views.}\label{fig:explain}
    \vspace{-1em}
\end{figure}

\section{Related Work}
\paragraph{Diffusion-based Image and Video Generation.}
Recent advancements have significantly expanded the capabilities of generating 2D images using diffusion models, as evidenced in several studies~\citep{dhariwal2021diffusion, nichol2021glide, podell2023sdxl, ramesh2022hierarchical, saharia2022photorealistic}. Notably, Stable Diffusion~\citep{rombach2022high} optimizes diffusion models (DMs) within the latent spaces of autoencoders, striking an effective balance between computational efficiency and high image quality. Beyond text conditioning, there is increasing emphasis on integrating additional control signals for more precise image generation~\citep{mou2024t2i, zhang2023adding}. For example, ControlNet~\citep{zhang2023adding} enhances the Stable Diffusion encoder to seamlessly incorporate these signals. Furthermore, the generation of multi-view images is gaining attention, with techniques like MVDiffusion~\citep{Tang2023mvdiffusion} or Geometry Guided Diffusion  \citep{song2023roomdreamertextdriven3dindoor} processing perspective images with a pre-trained diffusion model. Diffusion models are also extensively applied in video generation, as demonstrated by various recent works~\citep{ge2023preserve, ho2022imagen, wang2023modelscope, wu2023tune, wu2023lamp, zhou2022magicvideo}.
For instance, Imagen Video~\citep{ho2022imagen} utilizes a series of video diffusion models to generate videos from textual descriptions. Similarly, Make-A-Video~\citep{singer2022make} advances a diffusion-based text-to-image model to create videos without requiring paired text-video data.
MagicVideo~\citep{zhou2022magicvideo} employs frame-wise adaptors and a causal temporal attention module for text-to-video synthesis. Video Latent Diffusion Model (VLDM)~\citep{blattmann2023align} incorporates temporal layers into a 2D diffusion model to generate temporally coherent videos.

\vspace{-3mm}
\paragraph{3D/4D Large-scale Generation.}
In recent 3D computer vision, a large-scale scene is usually represented as implicit or explicit fields for its appearance~\citep{mildenhall2020nerf, kerbl20233d}, geometry~\citep{peng2020convolutional, wang2023alto, huang2023nksr}, and semantics~\citep{kerr2023lerf, zhou2024feature, qin2023langsplat}. We mainly discuss the 3D Gaussian Splatting (3DGS)~\citep{kerbl20233d} based generation here. Several works including DreamGaussian~\citep{tang2023dreamgaussian}, GaussianDreamer~\citep{yi2023gaussiandreamer}, GSGEN~\citep{chen2023text}, and CG3D~\citep{vilesov2023cg3d} employ 3DGS to generate diverse 3D objects and lay the foundations for compositionality, while LucidDreamer~\citep{chung2023luciddreamer}, Text2Immersion~\citep{ouyang2023text2immersion}, GALA3D~\citep{zhou2024gala3d}, RealmDreamer~\citep{shriram2024realmdreamer}, and DreamScene360~\citep{zhou2024dreamscene360} aim to generate static large-scale 3D scenes from text. Considering the current advancements in 3D generation, investigations into 4D generation using 3DGS representation have also been conducted. DreamGaussian4D~\citep{ren2023dreamgaussian4d} accomplishes 4D generation based on a reference image. AYG~\citep{ling2023align} equips 3DGS with dynamic capabilities through a deformation network for text-to-4D generation. Besides, Efficient4D~\citep{pan2024fast} and 4DGen~\citep{yin20234dgen} explore video-to-4D generation, and utilize SyncDreamer~\citep{liu2023syncdreamer} to produce multi-view images from input frames as pseudo ground truth for training a dynamic 3DGS. 
4K4D~\citep{xu20244k4d} is a high-resolution reconstruction technique that extends 3DGS to model complex human motion with detailed backgrounds while achieving real-time rendering speed.

\paragraph{Panoramic Representation.}
A panorama is an image that captures a wide, unbroken view of an area, typically encompassing a field of vision much wider than what a standard photo would cover, providing a more immersive representation of the subject. Recently, novel view synthesis using panoramic representation has been widely explored. For instance, PERF~\citep{wang2023perf} trains a panoramic neural radiance field from a single panorama to synthesize 360$^\circ$ novel views. 360Roam~\citep{huang2022360roam} proposed learning an omnidirectional neural radiance field and progressively estimating a 3D probabilistic occupancy map to speed up volume rendering. OmniNeRF~\citep{gu2022omni} introduced an end-to-end framework for training NeRF using only 360$^\circ$ RGB images and their approximate poses. PanoHDR-NeRF~\citep{gera2022casual} learns the full HDR radiance field from a low dynamic range (LDR) omnidirectional video by freely moving a standard camera around. In the realm of 3DGS, 360-GS~\citep{bai2024360} takes 4 panorama images and 2D room layouts as scene priors to reconstruct the panoramic Gaussian radiance field. DreamScene360~\citep{zhou2024dreamscene360} achieves text-to-3D Panoramic Gaussian Splatting by utilizing monocular depth priors to regularize the Gaussian optimization.

%% file: sec/method.tex
\section{Methodology}

Taking a single panoramic image as input, the goal of 4K4DGen is to generate a panoramic 4D environment capable of rendering novel views from arbitrary angles and at various timestamps. Our approach initially constructs a panoramic video and then elevates it into a series of 3D Gaussians, enabling efficient splatting for flexible rendering. Na\"ively animating projected perspective images, however, often results in unnatural motion and inconsistent animations. To overcome this, our method propose the denoising of projected spherical latents, ensuring consistent animation of the panoramic video from the original image, as detailed in Sec.~\ref{sec:animate_the_panorama}.

\begin{figure}[!ht]
  \centering
  \includegraphics[width=0.8\textwidth]{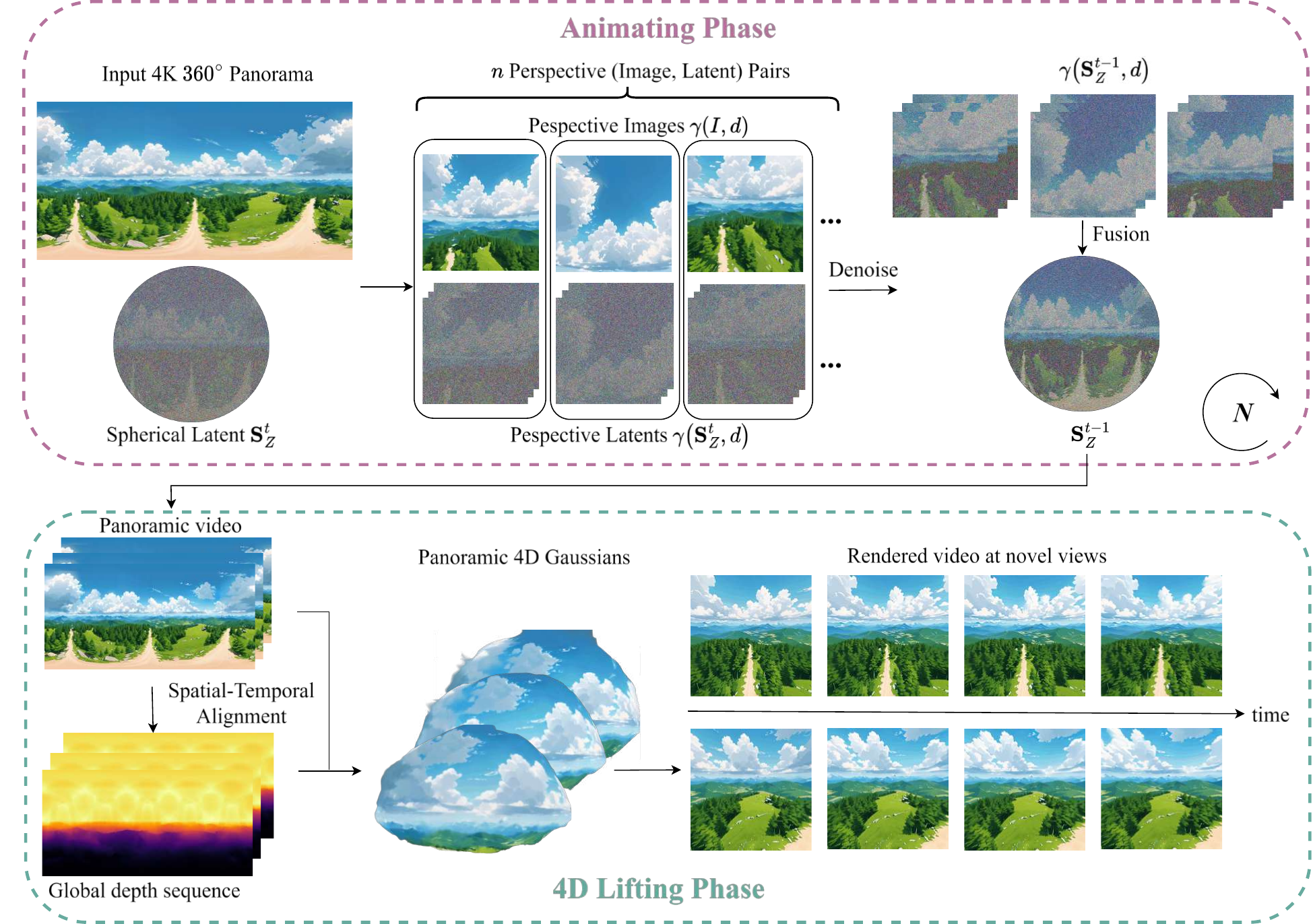}
    \caption{\textbf{Overall Pipeline.} Beginning with a static panorama as input, the \textbf{Animating Phase} generates a panoramic video by first mapping the panorama into a spherical latent space, followed by denoising within the perspective space, fusing back to the spherical latent space at each step, and finally transforming it into the panoramic space. In the \textbf{4D Lifting Phase}, a series of dynamic Gaussians is employed to lift the panoramic video into a 4D representation, ensuring both spatial and temporal consistency.}
  \label{fig:pipeline}
  \vspace{-3mm}
\end{figure}

Moreover, directly converting multiple perspective images from different timestamps into 4D frequently leads to degraded geometry and visible artifacts (see Sec.~\ref{ssec:exp_ablation}). We address this by applying spatial-temporal geometry fusion to lift the panoramic video, as described in Sec. \ref{sec:lifting}. The complete pipeline of 4K4DGen is illustrated in Fig. \ref{fig:pipeline}.

\subsection{Preliminaries}
\paragraph{Latent Diffusion Models (LDMs).}
LDMs ~\citep{rombach2021highresolution} consist of a forward procedure $q$ and a backward procedure $p$. The forward procedure gradually introduces noise into the initial latent code $x_0 \in \real^{h \times w \times c}$, where $x_0 = \mathcal{E}(I)$ is the latent code of image $I$ within the latent space of a VAE, denoted by $\mathcal{E}$. Given the latent code at step $t-1$, the $q$ procedure is described as $q(x_t|x_{t-1}) = \mathcal{N}(x_t; \sqrt{1-\beta_t} x_{t-1}, \beta_t \Mat{I})$. Conversely, the backward procedure $p$, aimed at progressively removing noise, is defined as $p_{\theta}(x_{t-1}|x_t) = \mathcal{N}(\mu_{\theta}(x_t, t), \Sigma_{\theta}(x_t, t))$. In practical applications, images are generated under the condition $y$, by progressively sampling from $x_T$ down to $x_0$. Recently, image-to-video (I2V) generation has been realized ~\citep{guo2023animatediff, dai2023animateanything} by extending the latent code with an additional frame dimension and performing decoding at each frame. The denoising procedure is succinctly represented as $x_{t-1} = \Phi(x_t, I)$, where $x_t, x_{t-1} \in \real^{l \times h \times w \times c}$ represent the sampled latent codes and $I$ the conditioning image.
Recently, image-to-video (I2V) generation has been achieved~\citep{guo2023animatediff, dai2023animateanything} by extending the latent code with an additional frame dimension and performing decoding at each frame. The denoising procedure is succinctly expressed as $x_{t-1} = \Phi(x_t, I)$, where $x_t, x_{t-1} \in \real^{l \times h \times w \times c}$ represent the sampled latent codes, and $I$ represents the conditioning image.

\paragraph{Omnidirectional Panoramic Representation.}
\label{sec:projection}
Panoramic images or videos, denoted as $I$, are typically represented using equirectangular projections, forming an $H \times W \times C$ matrix, where $H$ and $W$ denote the image resolution and $C$ represents the number of channels. While this format preserves the matrix structure, making it consistent with planar images captured by conventional cameras, it introduces distortions, especially noticeable near the polar regions of the projection. To mitigate these distortions, we adopt a spherical representation for panoramas, where pixel values are defined on a sphere
$\mathbb{S}^2 = \{\Mat{d} = (x, y, z) | x, y, z \in \real \wedge |\Mat{d}| = 1\}$. 
For a more precise definition of the projection, we represent matrix-like images using a mapping $\mathcal{E}_I: [-1, 1]^2 \rightarrow \real^C$, which normalizes the image coordinates into the range $[0,1]$. Thus, for any given pixel $\left(x, y\right) \in [-1,1]^2$, the corresponding pixel value is determined by $\mathcal{E}_I\left(x, y\right)$.
We define the spherical representation of panoramas using the field $\mathcal{S}_I: \mathbb{S}^2 \rightarrow \real^C$, where $\mathcal{S}_I\left(\Mat{d}\right)$ gives the pixel value at a given direction $\Mat{d} = (x, y, z)$. The relationship between the spherical and equirectangular representations is established through the following projection formula:
\begin{equation}
    \mathcal{S}_I\left(x,y,z\right) = \mathcal{E}_I\left(\frac{1}{\pi}\arccos\frac{y}{\sqrt{1-z^2}},\frac{2}{\pi}\arcsin z\right).
    \label{equ_proj}
\end{equation}
For perspective images, we define a virtual camera centered at the origin. The rays for each pixel are determined through ray casting, as described in ~\citep{mildenhall2020nerf}, where each ray $\Mat{d}$ is represented by $\Mat{r}(x, y, f, \Mat{u}, \Mat{s}, R)$. This representation takes into account the focal length $f$, the z-axis direction $\Mat{u}$, the image plane size $\Mat{s}$, and the camera's rotation along the z-axis $R$. Consequently, for a given panorama $I$, the perspective image $P$ can be projected using these camera parameters ($f, \Mat{u}, \Mat{s}, R$) as:
\begin{equation}
\mathcal{E}_P\left(x, y\right)= \mathcal{S}_I\circ\Mat r\left(x,y,f,\Mat u,\Mat s, R\right).
\label{pers_proj}
\end{equation}
In this paper, we fix the focal length $f$, the image plane size $\Mat{s}$, and the rotation $R$. We denote the process of projecting the panorama $I$ into a perspective image $i$, based on the camera's $z$-axis direction $\Mat{u}$, as $i = \gamma(I, \Mat{u})$.

\subsection{Inconsistent Perspective Animation}
Large-scale pre-trained 2D models have shown remarkable generative capabilities in creating images and videos, benefiting from vast multi-modal training data gathered from the Internet. However, acquiring high-quality 4D training data is considerably more challenging, and no current 4D dataset reaches the scale of those available for images and videos. Therefore, our approach aims to utilize the capabilities of video generative models to produce consistent panoramic 360$^\circ$ videos, which are then elevated to 4D. Nonetheless, the availability of panoramic videos is significantly more limited compared to planar perspective videos. Consequently, mainstream image-to-video (I2V) animation techniques may not perform optimally for panoramic formats, and the resolution of the videos remains constrained, as illustrated in Fig. \ref{fig:abl1} (b) and Tab. \ref{tab:abl-animating}. 
Alternatively, the animator can be applied to perspective images. but this introduces inconsistencies across different projected views, as depicted in Fig. \ref{fig:abl1} (c)

\subsection{Consistent Panoramic Animation}
\label{sec:animate_the_panorama}
Limited by the scarcity of 4D training data in panoramic format, and given that large diffusion models are primarily trained on planar perspective videos, directly applying 2D perspective denoisers presents challenges in generating seamless panoramic videos with proper equirectangular projection, due to inconsistent motion across different views and the domain gap between spherical and perspective spaces.
This constraint has driven us to develop a panoramic video generator in spherical space that leverages priors from general image-to-video (I2V) animation techniques, as shown in Fig.~\ref{fig:explain}.
Consequently, starting from a static input panorama, we animate it into a panoramic video, as demonstrated in the "Animating Phase" section of Fig. \ref{fig:pipeline}.

\paragraph{Spherical Latent Space.}  
To generate panoramic video from a static panorama, we build up the denoise-in-latent-space schema~\citep{an2023latent, blattmann2023stable, dai2023animateanything} in a spherical context. For general video generation, a noisy latent sample is progressively denoised using DDPM ~\citep{ho2020denoising}, conditioned on a static input image, and subsequently decoded into a video sequence by a pre-trained VAE decoder.
However, in 4K4DGen, unlike the method for generating perspective planar videos, both the latent code and the static panorama input are represented on spheres. We start with the initial panoramic latent code $S^T:\mathbb{S}^2\rightarrow \real^{L \times c}$, where $L$ denotes the number of video frames and $c$ the channels per frame.  A novel Panoramic Denoiser is then applied to generate the clean panoramic latent code $S^0$, conditioned on the static input panorama $I \in \real^{H \times W}$. Subsequently, the equirectangular projection, as introduced in Sec. \ref{sec:projection}, projects the clean panoramic latent code into the matrix-like latent code $Z^0 \in \real^{h \times w \times L \times c}$, with $h$ and $w$ representing the resolution of the latent code. Each $\ord{k}$ video frame $I^k$ in pixel space is decoded by the pre-trained VAE decoder as $I^k = \mathcal{D}(Z^0[:,:,k,:])$.

\paragraph{Build the Panoramic Denoiser.} 
We leverage a pre-trained perspective video generative model \citep{dai2023animateanything} to biuld our Panoramic Denoiser. 
This video generator takes a perspective image $i \in \real^{p_H \times p_W \times c}$ and an initial latent code $z^T \in \real^{p_h \times p_w \times (L \times c)}$ as inputs, 
progressively denoising the latent code $z^T$ to a clean state $z^0$ through a denoising function $z^{t-1} = \Phi(z^t, i)$.
Here, $p_h$ and $p_w$ represents the resolution of the latent code, $p_H$ and $p_W$ the resolution of the conditioning image, $c$ the number of channels, and $L$ the video length.
Our goal is to transform the initial noisy panoramic latent code $S^T$ into the clean state $S^0$, ensuring that each perspective view is appropriately animated while maintaining global consistency. The underlying intuition is that if each perspective view undergoes its respective denoising process, the perspective video will feature meaningful animation. Moreover, if two perspective views overlap, they will align with each other ~\citep{jimenez2023mixture, bar2023multidiffusion, lugmayr2022repaint} to produce a seamless global animation.

Given a static input panorama $I$ and an initial spherical latent code $S^0: \mathbb{S}^2 \rightarrow \real^{L \times c}$, we progressively remove noise employing a project-and-fuse procedure at each denoising step. Specifically, the spherical latent code at the $\ord{t}$ denoising step, $S^t: \mathbb{S}^2 \rightarrow \real^{L \times c}$, is projected into multiple perspective latent codes $\mathcal{Z}^t = \{z^t_1, z^t_2, \ldots, z^t_n\}$, where each $z^t_k = \gamma(S^t, \Mat{d}_k) \in \real^{p_h \times p_w \times (L \times c)}$ represents the $\ord{k}$ perspective latent code projected in the equirectangular format detailed in Sec. \ref{sec:projection}. Each perspective latent code is then denoised by one step using a pre-trained perspective denoiser, denoted as $z_k^{t-1} = \Phi(z^t_k, i_k)$, where $i_k = \gamma(I, \Mat{d}_k) \in \real^{p_H \times p_W \times c}$ is the perspective conditioning image projected from the panorama $I$. Subsequently, we optimize the spherical latent code $S^{t-1}: \mathbb{S}^2 \rightarrow \real^{L \times c}$ at step $t-1$ by fusing all the denoised perspective latent codes ${z_k^{t-1}}$. Formally, the denoising procedure at step $t$, denoted as $S^{t-1} = \Psi(S^t, I)$, encompasses the following operations:
\begin{equation}
\Psi\left(\mathcal{S}^{t}, I\right)=\operatornamewithlimits{argmin}_{\mathcal{S}} \mathbb{E}_{\Mat d\in \mathbb{S}^2} \|\gamma(\mathcal{S}, \Mat d)-\Phi\left(\gamma(\mathcal{S}^t, \Mat d),\gamma(I, \Mat d)\right)\|.
\end{equation}

\subsection{Dynamic Panoramic Lifting} \label{sec:lifting}
We define the panoramic video as $V = \{I^1, I^2, \ldots, I^L\}$, consisting of $L$ frames. The video is divided into overlapping perspective videos $\{v_0, v_1, \ldots, v_n\}$, each captured from specific camera directions $\{\Mat{d}_1, \ldots, \Mat{d}_n\}$, collectively encompassing the entire span of the panoramic video $V$. Subsequently, we estimate the geometry of the 4D scene by fusing the depth maps through spatial-temporal geometry alignment. Following this, we describe our methodology for 4D representation and the subsequent rendering procedure.

\paragraph{Supervision from Spatial-Temporal Geometry Alignment.}
To transition from 2D video to 3D space, we utilize a monocular depth estimator ~\citep{ranftl2021vision}, inspired by advancements in ~\citep{zhou2024dreamscene360}, to estimate the scene’s geometric structure. Nonetheless, depth maps generated for each frame and perspective might lack spatial and temporal consistency. To address this, we implement Spatial-Temporal Geometry Alignment using a pre-trained depth estimator $\Theta: \real^{h \times w \times 3} \rightarrow \real^{h \times w}$, applied to perspective images. Our objective is to amalgamate $n$ perspective depth maps $D^K_i = \Theta(\gamma(I^k, \Mat{d}_i))$ into a cohesive panoramic depth map $D^k$ for each frame $I^k$, ensuring spatial and temporal continuity. We express these depth maps as a spherical representation ${\mathcal{S}_D^1, \ldots, \mathcal{S}_D^L}$. For enhanced optimization, we assign $n$ scale factors $\alpha_i^k \in \real$ and shifting parameters $\beta_i^k \in \real^{h \times w}$ to each perspective depth map. The comprehensive depth map $D^k$ is then optimized jointly with these parameters $\alpha$ and $\beta$. The formal objective is structured as follows:
\begin{equation}
\centering
    \mathcal{S}_D^k = \operatornamewithlimits{argmin}_{\mathcal{S}}  \operatornamewithlimits{\mathbb{E}}_{i \in \left\{1,...n\right\}}  \lambda_{\rm{depth}} \mathcal{L}_{\rm{depth}} 
     +\lambda_{\rm{scale}} \mathcal{L}_{\rm{scale}} 
    + \lambda_{\rm{shift}} \mathcal{L}_{\rm{shift}}. 
    \label{eq:lift}
\end{equation}
where $\mathcal{L}_{\rm{depth}}=\|\operatorname{softplus}(\alpha_i^k)\Theta(\gamma(I^k, d_i))-\gamma(\mathcal{S}) + \beta_i^k\|$ is the depth supervision term, $\mathcal{L}_{\rm{scale}}=\|\alpha_i^k-\alpha_i^{k-1}\| +\|\operatorname{softplus}(\alpha_i^k)-1\|$ the regularize term for $\alpha$, and $\mathcal{L}_{\rm{shift}}=\mathcal{L}_{\rm{TV}}(\beta_i^k) + \|\beta_i^k-\beta_i^{K-1}\|$ the regularize term for $\beta$ where $\mathcal{L}_{\rm{TV}}$ is the TV regularization.

\paragraph{4D Representation and Rendering.}
We represent and render the dynamic scene using $T$ sets of 3D Gaussians. Each set, corresponding to a specific timestamp $t$, is denoted as $G_t = \{\left(\Mat{p}_t^i, \Mat{q}_t^i, \Mat{s}_t^i, \Mat{c}_t^i, o_t^i\right) | i = 1, \ldots, n\}$. This definition aligns with the methods described in ~\citep{bahmani20234d}, which also provides a fast rasterizer for rendering images based on these Gaussian sets and given camera parameters. Consistent with Sec. \ref{sec:projection}, while the camera intrinsics remain fixed, we parameterize the camera extrinsics through a position $\Mat{p} \in \real^3$ and an orientation $\Mat{d} \in \mathbb{S}^2$. The training process is structured in two stages: initially, we directly supervise the 3D Gaussians using the panoramic videos. Let $\mathcal{R}(G, \Mat{p}, \Mat{d})$ represent the rasterized image from Gaussian set $G$, utilizing camera extrinsics $\Mat{p} = 0$ and camera direction $\Mat{d}$. Let $I_t$ denote the $\ord{t}$ frame of the panoramic video. We optimize the $\ord{t}$ Gaussian set $G_t$ using the following objective:
\begin{equation}
\label{eq:L4d}
    \mathcal{L} = 
    \lambda_{\rm{rgb}} \mathcal{L}_{\rm{rgb}}+
    \lambda_{\rm{temporal}} \mathcal{L}_{\rm{temporal}}+
    \lambda_{\rm{sem}} \mathcal{L}_{\rm{sem}}+
    \lambda_{\rm{geo}} \mathcal{L}_{\rm{geo}}
\end{equation}
where the RGB supervision term $\mathcal{L}_{\rm{rgb}}=\lambda \mathcal{L}_1 + (1-\lambda) \mathcal{L}_{\rm{SSIM}}$ is the same as 3D-GS ~\citep{kerbl20233d}, and the temporal regularize term $\mathcal{L}_{\rm{temporal}}$ written as:

\begin{equation}
    \mathcal{L}_{\rm{temporal}}=\sum\limits_{i=1}^{n}\|\mathcal{R}(G_t, \Mat 0, \Mat d_i)-\mathcal{R}(G_{t-1}, \Mat 0, \Mat d_i))\|
\end{equation}

Then, we adopt the distillation loss and geometric regularization used in ~\citep{zhou2024dreamscene360}, the distillation loss is defined as follows: $\mathcal{L}_{sem} = 1-\operatorname{cos}\left\langle \operatorname{CLS}(\mathcal{R}(G_t, \Mat 0, \Mat d_i)),  \operatorname{CLS}(\mathcal{R}(G_t,\Mat \delta_p,\Mat d_i))\right\rangle$,
where $\Mat\delta_p\in[-\alpha, \alpha]^3$ is the disturbing vector, $\operatorname{CLS}(\cdot)$ the feature extractor such as DINO~\citep{oquab2023dinov2}, and $\operatorname{cos}\langle\cdot,\cdot\rangle$ the $\operatorname{cos}$ value of two vectors. The geometric regularization is defined as follows: $\mathcal{L}_{geo} = 1 - \frac{\operatorname{Cov}(\mathcal{R}_D(G_t, \Mat 0, \Mat d_i), \Theta(\gamma(I, \Mat d_i)))}{\sqrt{\operatorname{Var}(\mathcal{R}_D(G_t, \Mat 0, \Mat d_i))\operatorname{Var}(\Theta(\gamma(I, \Mat d_i)))}}$, where $\mathcal{R}_D$ is the rendered depth, $\operatorname{Cov}(\cdot, \cdot)$ the covariance, and $\operatorname{Var}(\cdot)$ the variance.

\section{Experiments}

\begin{figure}[tb]
  \centering
  \includegraphics[width=0.9\textwidth]{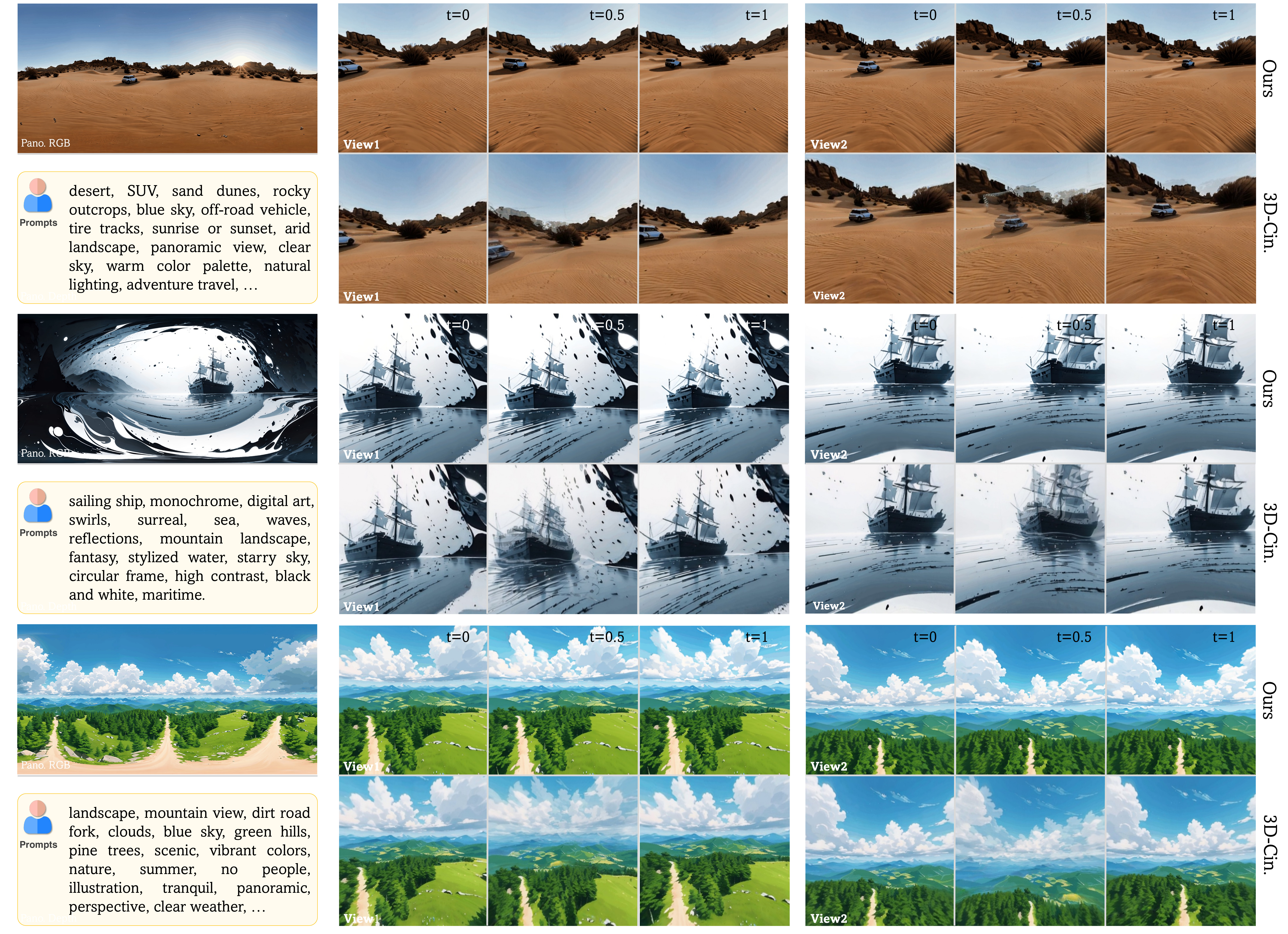}
 \caption{\textbf{Comparison between 4K4DGen and 3D-Cinemagraphy.} We present the input static panorama (Pano RGB), the corresponding text prompts, and the rendered results from different views and at various timestamps. 4K4DGen (Ours) effectively generates 4D scenes that are both spatially and temporally consistent, while 3D-Cinemagraphy (3D-Cin.) suffers from ghosting artifacts in the middle frames.}
  \label{fig:com}
\end{figure}

%% file: sec/exps.tex
\vspace{-3mm}
\subsection{Experimental Settings}
\label{sec:experimental_settings}
\paragraph{Implementation Details.} 
\vspace{-2mm}

For perspective images, we uniformly select 20 directions $\Mat u$ on the sphere $\mathbb{S}^2$ as the z-axis of 20 cameras. In each experiment, the image plane size $\Mat s$ is set at $0.6 \times 0.6$, with a focal length $f = 0.6$ and a resolution of $512 \times 512$. Rotation along the z-axis is kept at zero for all cameras, ensuring that the up-axis for the $\ord i$ camera aligns with the $(O, \Mat u_i, \Mat z)$ plane.
During the animating phase, we utilize the perspective denoiser $\Phi$, instantiated as the Animate-anything model ~\citep{dai2023animateanything}, which fine-tunes the SVD model ~\citep{blattmann2023stable}. In the Spatial-Temporal Geometric Alignment stage of the lifting phase, the depth estimator $\Theta$ is implemented using MiDaS ~\citep{ranftl2021vision, birkl2023midas}. 
All experiments are executed on a single NVIDIA A100 GPU with 80 GB RAM.

\vspace{-3mm}
\paragraph{Evaluation.} 
As there is no ground truth 4D scene data available, we render videos at specific test camera poses from the synthesized 4D representation and employ non-reference video/image quality assessment methods for quantitative evaluation of our approach. For the test views, we select random cameras with $\Mat p=0$ as part of our testing camera set. We then introduce disturbances as described in Sec. \ref{sec:lifting}, applying a disturbance factor of $\alpha=0.05$ at these selected views.
\underline{Datasets.} The task of generating 4D panoramas from static panoramas is new, and thus, no pre-existing datasets are available. In line with previous large-scale scene generation works \citep{zhou2024dreamscene360, yu2024wonderworld}, we evaluate our methodology using a dataset of 16 panoramas generated by text-to-panorama diffusion models.
\underline{Baselines.} Current SDS-based methods ~\citep{wu20234d, zhao2023animate124} are limited to generating object-centered assets and do not support outward-facing scene generation. We compare our method with the optical-flow-based 3D dynamic image technique, 3D-Cinemagraphy (3D-Cin.)~\citep{li2023_3dcinemagraphy} (both the ``circle'' and ``zoom-in'' mode), by inputting the static panorama and projecting the output onto perspective images. 
\underline{Metrics.} It is challenging to evaluate the visual quality without a ground-truth reference. 
We assess the rendered perspective videos based on both frame and video visual quality.
For frame quality, we use distribution-based metrics such as FID \citep{heusel2017gans} and KID \citep{binkowski2018demystifying}, which calculate the distance between generated frames and the corresponding perspective images projected from the static panoramas.
We also employ the LLM-based visual scorer Q-Align \citep{wu2023qalign} to evaluate the quality of individual frames.
For video quality, we use the Q-Align video model as the quality scorer. Additionally, we conduct user studies to further evaluate the results. 
In this paper, there are two types of user studies: (1) User Choice (UC), where participants are asked to compare and select the best video from candidates generated by different methods, and (2) User Agreement (UA), where participants assess whether specific properties are present in the videos generated by a particular approach.

\begin{table}[!t]
\caption{\textbf{Comparison with 3D-Cinemagraphy.} 
The IQ, IA, and VQ models represent the image quality scorer, image aesthetic scorer, and video quality scorer, respectively, within the Q-Align assessment framework. Our method, 4K4DGen, consistently achieves superior performance in both image and video quality across these metrics. Furthermore, the majority of participants in our user study rated 4K4DGen as the best in terms of visual quality.} 
\vspace{-3mm}
\label{table:comparison}
 \centering
 \resizebox{0.99\textwidth}{!}{
\begin{tabular}{c|ccccccc}
\toprule[1.2pt]
\bf{Method}  & FID  $\downarrow$ & KID ($\times 10^{-2}$) $\downarrow$  & 
 Q-Align (IQ) $\uparrow$ & Q-Align (IA) $\uparrow$ & Q-Align (VQ) $\uparrow$ & Quality (UC)  $\uparrow$ \\
\midrule[1.2pt]
3D-Cinemagraphy (zoom-in) & 57.05 & 1.4  & 0.47 & 0.38 & 0.57 & 7\%\\

3D-Cinemagraphy (circle) &    57.78 & 1.3  & 0.48 & 0.40 & 0.58  & 12\% \\

\cellcolor{mygray}4K4DGen& \cellcolor{mygray}\textbf{16.59} & \cellcolor{mygray}\textbf{0.2} & \cellcolor{mygray}\textbf{0.66} & \cellcolor{mygray}\textbf{0.44} & \cellcolor{mygray}\textbf{0.62}
& \cellcolor{mygray} \textbf{81\%} \\
 \bottomrule[1.2pt]
\end{tabular}
}
\vspace{-4mm}
\end{table}

\subsection{Results}
\vspace{-2mm}
\paragraph{Quantitative Results.}
We also show the qualitative comparison between 4K4DGen and 3D-Cinemagraphy \citep{li20233d} in Tab. \ref{table:comparison}, considering both frame and video quality. In terms of frame quality, 4K4DGen achieves significantly better performance on both distribution-based metrics and LLM-based metrics. In terms of video quality, 4K4DGen achieves better Q-Align score, and is selected as the method with best visual quality by \textbf{81\%} of users in our study.

\paragraph{Qualitative Results.} 
We present a qualitative comparison between 4K4DGen and 3D-Cinemagraphy (3D-Cin.). Since the performance of 3D-Cin. is similar under the "circle" and "zoomin" settings in Tab. \ref{table:comparison}, we use the "circle" setting to represent 3D-Cin. in Fig. \ref{fig:com}. As shown in the figure, 4K4DGen produces high-quality perspective videos that maintain consistency across both time and views, whereas 3D-Cin. struggles with generating ghosting artifacts in the middle frames.

\begin{figure}[b]
  \centering
  \includegraphics[width=0.9\textwidth]{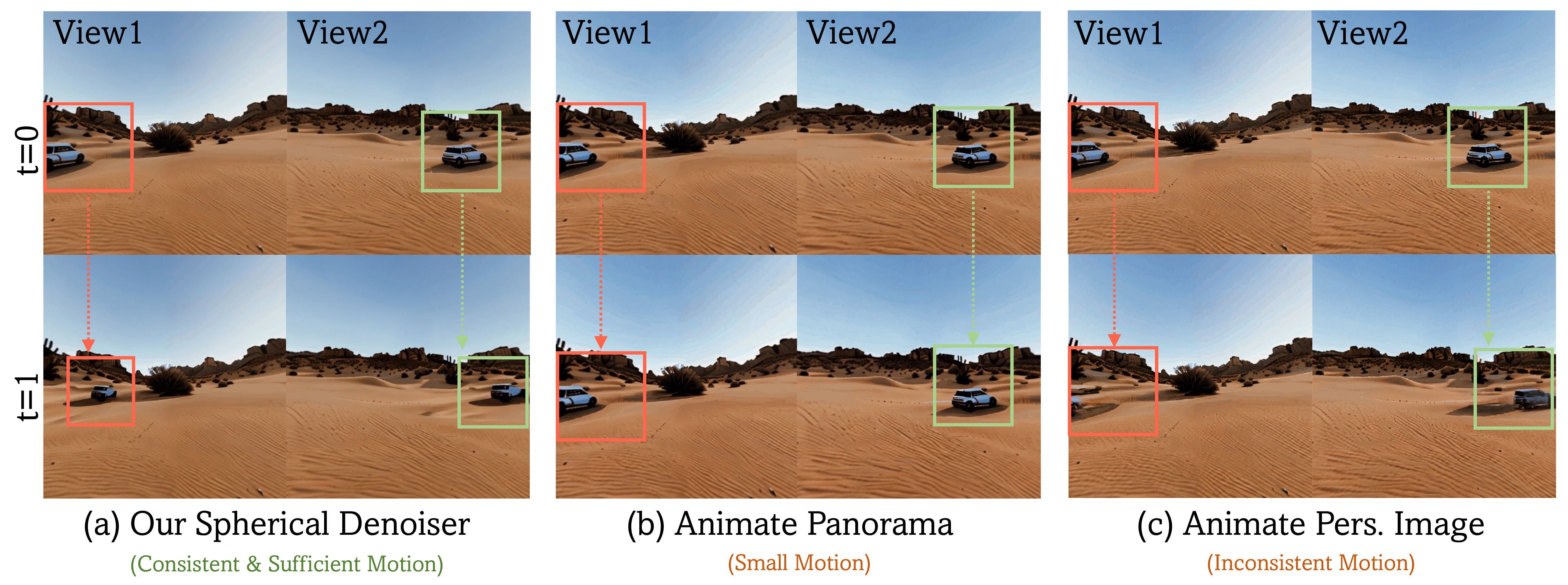}
  \caption{\textbf{Comparison to Different Animators}: Animators trained primarily on perspective images tend to produce limited motion when applied to panoramas, and the resolution may be limited. On the other hand, animating perspective images individually can lead to inconsistencies between overlapping views.}
  \label{fig:abl1}
\end{figure}

\vspace{-3mm}

\subsection{Ablation Studies}
\label{ssec:exp_ablation}
We conduct ablation studies for both the animating and lifting phases of our methodology. In the animating phase, we highlight the importance of our spherical denoise strategy by replacing it with two basic animation techniques. In the lifting phase, we analyze the impact of excluding the Spatial-Temporal Geometry Alignment process and the temporal loss during the optimization of 4D representations.
\paragraph{Animating Phase.} 
To animate the panorama into a panoramic video, a straightforward approach is to apply animators directly to the entire panorama. However, we observed that this strategy often results in minor motion, as shown in Fig. \ref{fig:abl1} (b) and Tab. \ref{tab:abl-animating} (Animate Pano.). This issue arises due to two main reasons: (1) animators are typically trained on perspective images with a narrow field of view (FoV), whereas panoramas have a 360$^\circ$ FoV with specific distortions under the equirectangular projection; (2) our panorama is high-resolution (4K), which exceeds the training distribution of most 2D animators and can easily cause out-of-memory issues, even with an 80GB VRAM graphics card. Thus the panoramas have to be down-sampled to a lower resolution (2K), causing a loss of details.
Thus, we seek to animate on perspective views. Applying the animator to perspective views offers benefits such as reduced distortion and domain-appropriate input for the animator, allowing for smooth animation of high-resolution panoramas. However, animating perspective images separately can introduce inconsistencies between overlapping perspective views, as illustrated in Fig. \ref{fig:abl1} (c) and Tab. \ref{tab:abl-animating} (Animate Pers.). To resolve this challenge, we propose simultaneously denoising all perspective views and fusing them at each denoising step, in the spherical latent spaace, which capitalizes on the benefits of animating perspective views while ensuring cross-view consistency. The results are displayed in Fig. \ref{fig:abl1} (a) and Tab. \ref{tab:abl-animating} (Ours).

\begin{figure}[!t]
  \centering
  \includegraphics[width=0.9\textwidth]{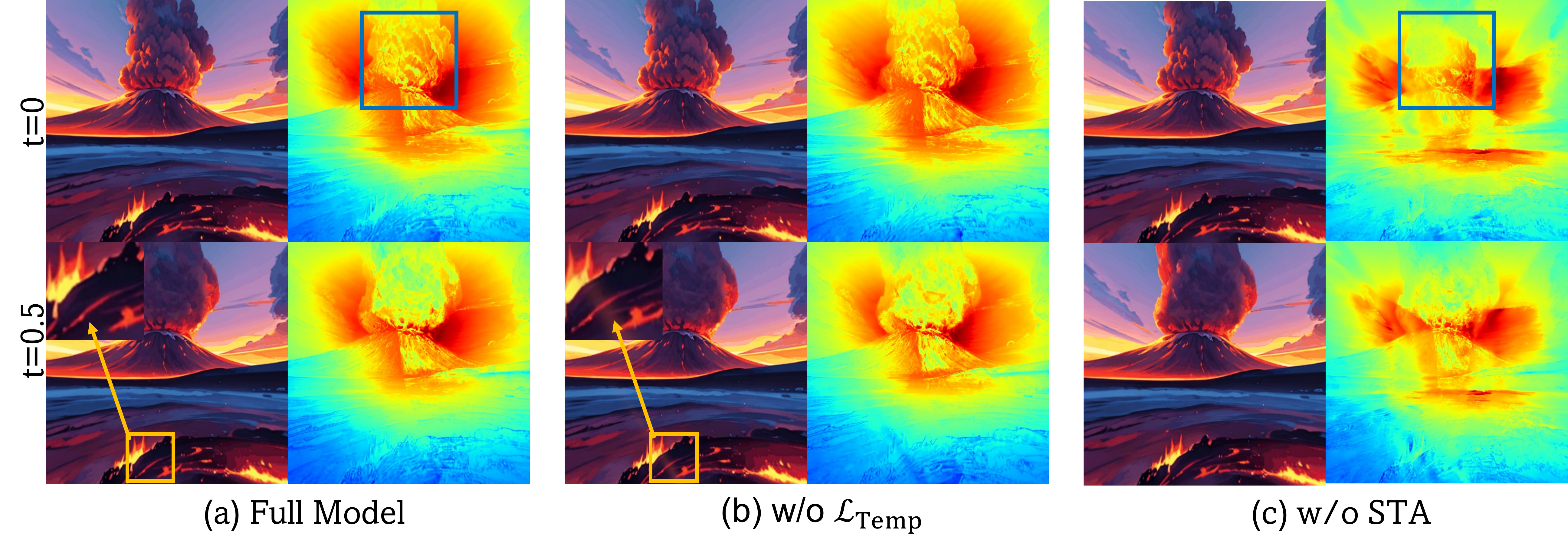}
  \vspace{-3mm}
  \caption{\textbf{Ablation of the Lifting Phase.} Omitting temporal regularization during the optimization of 3D Gaussians results in the appearance of artifacts. The absence of Spatial-Temporal Geometry Alignment causes the degradation of geometric structures.}
  \label{fig:abl2}
  \vspace{-4mm}
\end{figure}

\vspace{-2mm}
\paragraph{Lifting Phase.} 
We conduct ablation studies on the Spatial-Temporal Geometry Alignment (STA) module and the temporal loss during the lifting phase. Our findings indicate that removing the STA module leads to a degradation in geometric quality, as shown in Fig. \ref{fig:abl2} (c). Additionally, omitting the temporal loss introduces artifacts in certain frames, potentially resulting in flickering, as demonstrated in Fig. \ref{fig:abl2} (b).

\begin{table*}[!htb]
    \centering
    \caption{\textbf{Different Animation Strategies in the Animating Phase}. While the three strategies achieves similar visual quality (Q-Align), animating the entire panorama results in minor motion and reduced resolution (first row). Conversely, animating from perspective views leads to inconsistencies across different views (second row). This is supported by the ``user agreement (UA)'' study, where \textbf{70\%} of participants identified our method as view-consistent, in contrast to only \textbf{33\%} who consider animation from perspective views to be view-consistent.} 
    \label{tab:abl-animating}
    \renewcommand\arraystretch{1.3}
    \resizebox{0.8\textwidth}{!}{
        \begin{tabular}{l|c|cccccccc}
            \toprule[1.2pt]
\bf{Animater} & Max Pano. Res.    & Q-Aline(VQ) $\uparrow$  &  Motion$\uparrow$ &View-consistency (UA)$\uparrow$ & \\
\midrule[1.2pt]

Animate Pano. & $2048\times1024$   & 0.82 & 0.30& -\\
Animate Pers. & $4096\times2048$  & 0.64  & 0.91 &33\%\\
\cellcolor{mygray}Ours & \cellcolor{mygray}$4096\times2048$   & \cellcolor{mygray}\textbf{0.85} &   \cellcolor{mygray}\textbf{1.27} &\cellcolor{mygray}  \textbf{70\%}\\         
\bottomrule[1.2pt]
\end{tabular}
\vspace{-2mm}
}
\end{table*}

%% file: sec/conclusion.tex
\section{Conclusion}
\label{sec:limitation}
\vspace{-2mm}
\paragraph{Conclusion.}
We have proposed a novel framework \textbf{4K4DGen}, allowing users to create high-quality 4K  panoramic 4D content using text prompts, which delivers immersive virtual touring experiences.
To achieve panorama-to-4D even without high-quality 4D training data, we integrate generic 2D prior models into the panoramic domain. Our approach involves a two-stage pipeline: initially generating panoramic videos using a Panoramic Denoiser, followed by 4D elevation through a Spatial-Temporal Geometry Alignment mechanism to ensure spatial coherence and temporal continuity.

\vspace{-1mm}
\paragraph{Limitation.}
First, the quality of temporal animation in the generated 4D environment mainly relies on the ability of the pre-trained I2V model. Future improvements could include the integration of a more advanced 2D animator.
Second, since our method ensures spatial and temporal continuity during the 4D elevation phase, it is currently unable to synthesize significant changes in the environment, such as the appearance of glowing fireflies or changing weather conditions.
Third, the high-resolution and time-dependent representation of the generated 4D environment necessitates substantial storage capacity, which could be optimized in future work using techniques such as model distillation and pruning.

%% file: sec/supp.tex
\clearpage
\appendix
\section{Appendix}

Due to space constraints in the main draft, we include supplementary details and experimental results in the appendix. Specifically, in Sec. \ref{sec:sup-acq}
, we provide details about the acquisition process for the static panoramas used in our experiments. In Sec. \ref{sec:sup-imp}, we offer further explanation of the implementation for both the animation and lifting phases. Finally, in Sec. \ref{sec:sup-res}, we describe the experimental setup and present additional results.

\section{Acquisition  of Panoramas}
\label{sec:sup-acq}
The static panoramas used in the dataset of the main draft are generated by a text-to-panorama diffusion model, fine-tuned from stable diffusion ~\citep{rombach2021highresolution} on SUN360. Similar to ~\citep{feng2023diffusion360}, this model follows three steps: circular blending, super-resolution, and refinement. The panoramas are initially at a resolution of $6144 \times 3072$ and then down-sampled to $4096 \times 2048$ using the bi-linear interpolation.

\section{Implementation Details}
\label{sec:sup-imp}
In this section, we introduce the implementation details of the panoramic animator and the 4D lifting procedure. 
\paragraph{Implementation of Spherical Representing}
For the spherical representation, the continuous spherical mapping $\mathcal{S}_I: \mathbb{S}^2 \rightarrow \real^C$ is instantiate as discrete point set $\mathcal{P} = \{p_i\}$, which is uniformly sampled from the sphere $\mathcal{S}_I$.
We first initialize a icosahedron with 20 triangle faces $\{f_i|i=1,\cdots, 20\}$ to approximate a real sphere $\mathbb{S}^2$. Then we uniformly sample a point set $P_i$ on each face $f_i$ and union all the point sets together as $\hat {\mathcal{P}} = \cup _{i=1}^{20}P_i$. We then obtain the discrete point set $\mathcal{P}$ by projecting $\hat {\mathcal{P}}$ onto the sphere $\mathbb{S}^2$ by $\mathcal{P} = \{{p_i}/{\|p_i\|}~|~ p_i \in \hat {\mathcal{P}}\}$.

\begin{figure}[ht]
  \centering
  \includegraphics[width=\textwidth]{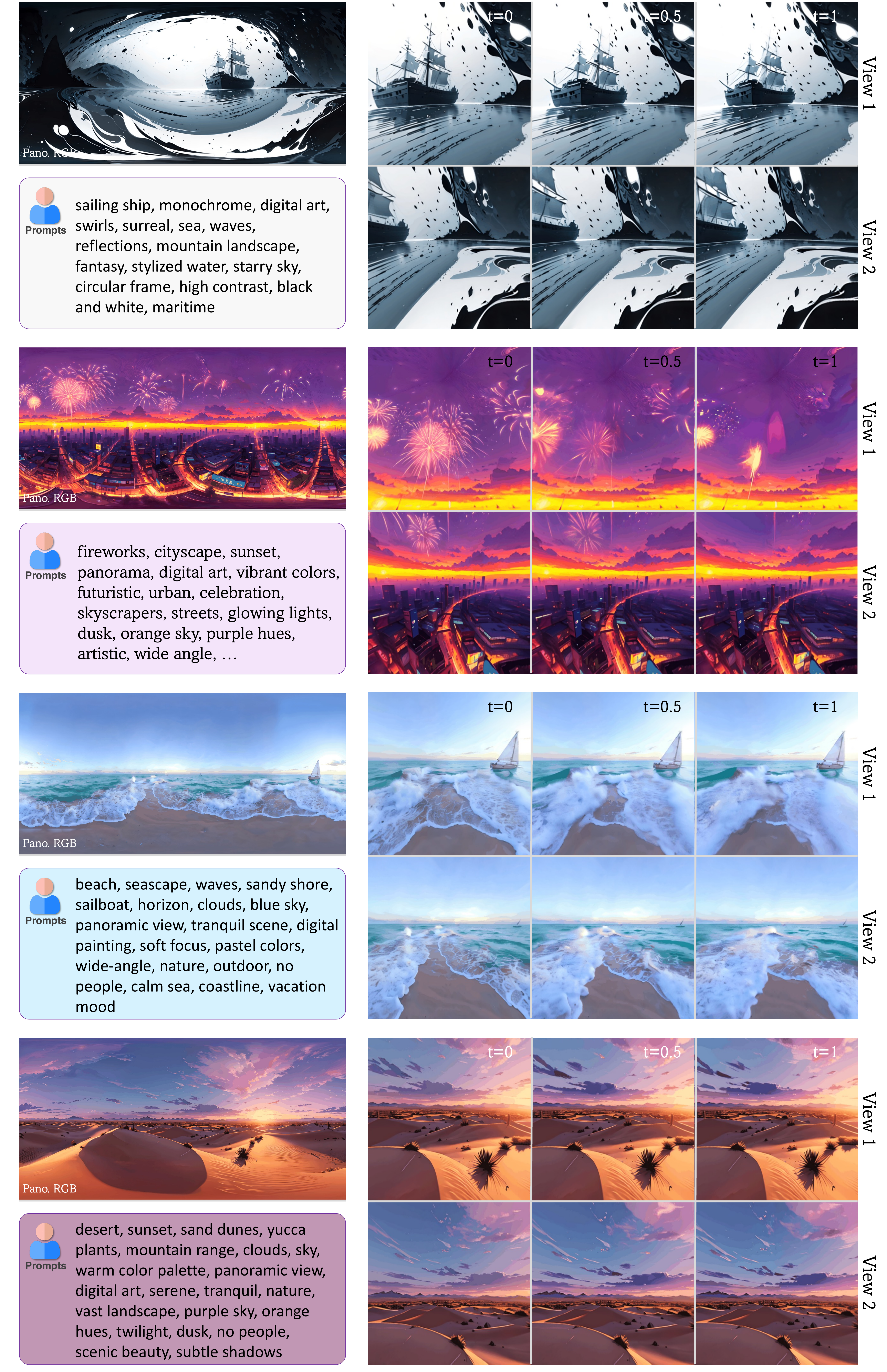}
  \caption{\textbf{Visualizations}: We provide more visual results. For each shown case we provide the input panorama, corresponding text prompt, and the rendering from two perspective views.}
  \label{fig:supp-vis}
\end{figure}

\paragraph{Panoramic Animation Phase}
For the Panoramic Animator, we set the video length $L = 14$, the channel number $c = 9$, the latent code size $(h, w) = \frac{1}{8} (H, W)$, the perspective image size $p_H = p_W = \frac{1}{4}W$. The sphere is uniformly divided into 20 perspective views, each with 80$^\circ$ FOV. For the denoiser, the max denoising step is $25$. For the continuous optimization in Eq. 3, we use a close form, where each latent vector at each point on the sphere is the average of the latent vectors of the corresponding pixel on the perspective views that overlap it. The perspective denoiser is initiated as Animate-Anything \cite{dai2023animateanything}. The masks required by the denoiser are given by bounding boxes defined by user clicks.

\paragraph{Dynamic Panoramic Lifting Phase} 
In the lifting phase, similar to the animation phase, we choose the perspective view number $n=20$, each with 80$^\circ$ FOV. Each perspective view has a square shape, $P_H = P_W = \frac{1}{4} W$, where $W$ is the width of the original static panorama. In the Spatial-Temporal Geometric Alignment stage, the depth estimator $\Theta$ is implemented using MiDaS ~\citep{ranftl2021vision, birkl2023midas}. The depth map from the perspective image is scaled according to the projection of the unit-length ray direction onto the camera orientation $\Mat d$. Formally, if the pre-scaled depth is $d$ at point $p\in \hat {\mathcal{P}}$ introduced above, the scaled depth should be $d / \|p\|$.
Additionally, for scenes without distinct boundaries, such as the sky, depth values for distant elements are assigned a finite value to support optimization.

\paragraph{Optimization Details}
The hyper-parameters for optimization are set as follows: $\lambda_{\rm{depth}}=1, \lambda_{\rm{scale}}=0.1, \lambda_{\rm{shift}}=0.01$. We conduct Spatial-Temporal Geometry Alignment optimization over 3000 iterations, with $\lambda_{\rm{scale}}$ and $\lambda_{\rm{shift}}$ set to zero during the first 1500 iterations.
For the 4D representation training stage, Gaussian parameters are optimized over 10000 iterations for each time stamp $t$. The hyper-parameters for this stage are defined as $\lambda_{\rm{rgb}}=1, \lambda_{\rm{temporal}}=\lambda_{\rm{sem}}=\lambda_{\rm{geo}}=0.05$, and the disturbance vector range $\alpha$ is varied at $0.05$, $0.1$, and $0.2$ during the $5400$, $6600$, and $9000$ iterations, respectively.

\section{Experimental Details}
\label{sec:sup-res}
\subsection{User Study Details}
We conducted two user studies, gathering a total of 84 questionnaires from 42 users. For the ``Quality (UC)''column in Tab. \ref{table:comparison}, we collected 42 questionnaires, each containing eight questions. 
Each question asked users to choose the bests video in term of visual quality from the perspective videos provided by different models. The user choice (UC) score of a method is the percentage of times the method’s video was selected as the best one, out of a total of 336 questions. Thus, the UC scores for all methods sum to 100\%.
For the ``View-Consistency (UA)'' column in Tab. \ref{tab:abl-animating}, we collected another 42 questionnaires, with each questionnaire containing eight questions. 
Each question presented two videos from different views, both generated by the same method, and users were asked to determine whether the two videos were view-consistent. The user agreement (UA) score is the percentage of video pairs marked as view-consistent out of all the video pairs generated by the method. The UA scores do not necessarily sum to 100\%.
In the UC column of Tab. \ref{table:comparison}, among the 336 questions, users selected 4K4DGen 272 times, 3D-Cin. (circle) 40 times, and 3D-Cin. (zoomin) 24 times. In the UA column of Tab. \ref{tab:abl-animating}, 118 out of 168 video pairs generated by ``Our'' were marked as consistent, while 56 out of 168 pairs from ``Animate Pers'' were considered consistent.

\subsection{More Results}
We provide additional qualitative results in Fig. \ref{fig:supp-vis}. Furthermore, we highly recommend viewing the video renderings of 4K4DGen and comparisons to baseline models in the supplementary static HTML page for a more comprehensive and immersive experience.

\section{Ethics and Reproducibility Statement} 
\paragraph{Ethics Statement.} Our research enables the generation of 4D digital scenes from a single panoramic image, which is advantageous for various applications such as AR/VR, movie production, and video games. This technology distinctly excels in creating high-resolution 4D scenes up to 4K, significantly enhancing user experiences. However, there is potential for misuse in the creation of deceptive content or privacy violations, which contradicts our ethical intentions. These risks can be mitigated through a combination of regulatory and technical strategies, such as watermarking.
\paragraph{Reproducibility.} We provide sufficient implementation details to reproduce our methodology in Sec. \ref{sec:sup-imp}, including the details of spherical denoiser, panoramic animator, dynamic panoramic lifting, etc. Furthermore, we will release the code in the future.